%% file: main.tex
\documentclass[letterpaper, 10 pt, conference]{ieeeconf}  
\IEEEoverridecommandlockouts                              
\let\labelindent\relax
\usepackage{enumitem,amssymb}
\newlist{todolist}{itemize}{2}
\setlist[todolist]{label=$\square$}

\usepackage[usenames, dvipsnames]{color}
\usepackage[noadjust]{cite}
\usepackage{booktabs}
\usepackage{graphicx}
\usepackage{caption}
\usepackage{subcaption}
\usepackage{float}
\usepackage{amsmath}
\usepackage{algorithmicx}
\usepackage{subfiles}
\usepackage{multirow}
\usepackage{mathrsfs}
\graphicspath{{images}{../images/}}
\usepackage{graphicx}
\usepackage[utf8]{inputenc}
\usepackage{amsmath}
\usepackage[hidelinks]{hyperref}
\usepackage{pifont}
\usepackage{amssymb}

\input{commands.tex}

\usepackage[usenames,dvipsnames,table]{xcolor}
\usepackage{soul,xcolor}
\usepackage{siunitx}
\usepackage[export]{adjustbox}
\usepackage{multicol}
\usepackage{tikz}
\definecolor{taga}{RGB}{0,0,255}
\definecolor{tagb}{RGB}{0,67,169}
\definecolor{tagc}{RGB}{255,101,0}
\definecolor{tagd}{RGB}{255,0,0}
\definecolor{route}{RGB}{0,248,0}

\title{\LARGE \bf
TLCFuse: Temporal Multi-Modality Fusion Towards Occlusion-Aware Semantic Segmentation
}

\author{Gustavo Salazar-Gomez$^\dagger$$^{1}$, Wenqian Liu$^\dagger$$^{1}$, Manuel Diaz-Zapata$^{1,2}$, David Sierra-Gonzalez$^{3}$, Christian Laugier$^{1}$\\  
\thanks{$^\dagger$~G. Salazar-Gomez and W. Liu contributed equally to this work.}
\thanks{This work was supported by Toyota Motor Europe.}
\thanks{$^{1}$The authors are with Inria, Univ. Grenoble-Alpes, Grenoble, France {\tt\small\{firstname.lastname\}@inria.fr}}
\thanks{$^{2}$ Manuel Diaz-Zapata is with CITILab, INSA Lyon.}
\thanks{$^{3}$ This work was conducted when David Sierra-Gonzalez was affiliated with Inria.}
}

\begin{document}
\setstcolor{red}

\maketitle
\thispagestyle{empty}
\pagestyle{empty}
\setlength{\belowdisplayskip}{2pt}
\widowpenalty10000
\clubpenalty10000
\addtolength{\abovedisplayskip}{-5pt}

\begin{abstract}
In autonomous driving, addressing occlusion scenarios is crucial yet challenging. Robust surrounding perception is essential for handling occlusions and aiding navigation. State-of-the-art models fuse LiDAR and Camera data to produce impressive perception results, but detecting occluded objects remains challenging. 
In this paper, we emphasize the crucial role of temporal cues in reinforcing resilience against occlusions in the bird's eye view (BEV) semantic grid segmentation task.
We proposed a novel architecture that enables the processing of temporal multi-step inputs, where the input at each time step comprises the spatial information encoded from fusing LiDAR and camera sensor readings. 
We experimented on the real-world nuScenes dataset and our results outperformed other baselines, with particularly large differences when evaluating on occluded and partially-occluded vehicles. 
Additionally, we applied the proposed model to downstream tasks, such as multi-step BEV prediction and trajectory forecasting of the ego-vehicle. The qualitative results obtained from these tasks underscore the adaptability and effectiveness of our proposed approach.

\end{abstract}


\input{sections/introduction}
\input{sections/related_work}
\input{sections/approach}

\input{sections/experiments}
\input{sections/results}
\input{sections/conclusion}
\input{sections/acknowledgement}




%
%

\bibliographystyle{IEEEtran}
\bibliography{main}%

\end{document}

%% file: commands.tex
\newcommand \inR {\mathbb{R}}

\def\beq{\begin{equation}}
\def\eeq{\end{equation}}

%% file: sections/introduction.tex

\section{Introduction} \label{sec:introduction}

In the realm of autonomous driving, the ability to navigate complex and dynamic environments is contingent on a vehicle's capacity to perceive its surroundings accurately. 
However, there are instances when objects become partially or entirely obstructed by other elements, causing them to vanish from the ego-vehicle's perspective. Developing robust perception methods to deal with these occluded objects is crucial yet challenging, as it directly impacts the safety, efficiency, and reliability of navigation for autonomous driving.

One notable advancement in perception is the fusion of multiple sensor modalities, which leads to impressive results in object detection \cite{dsgn,deep-cont-fusion,li2022deepfusion,wang2021pointaugmenting,mv3d}, bird's eye view (BEV) semantic grid segmentation\cite{bevfusion,hendy2020fishing,salazar2022transfusegrid,diaz2023laptnet} and the other related tasks. Nevertheless, these state-of-the-art approaches still often struggle to detect occluded objects effectively and the research towards occlusion-aware models remains limited\cite{cross-view-transformers,bartoccioni2022lara,salazar2022transfusegrid,bevformer,pan2023baeformer}.

\begin{figure}[t!]
	\centering
	    \includegraphics[trim={0 0 0 0},clip,width=1.0\columnwidth]{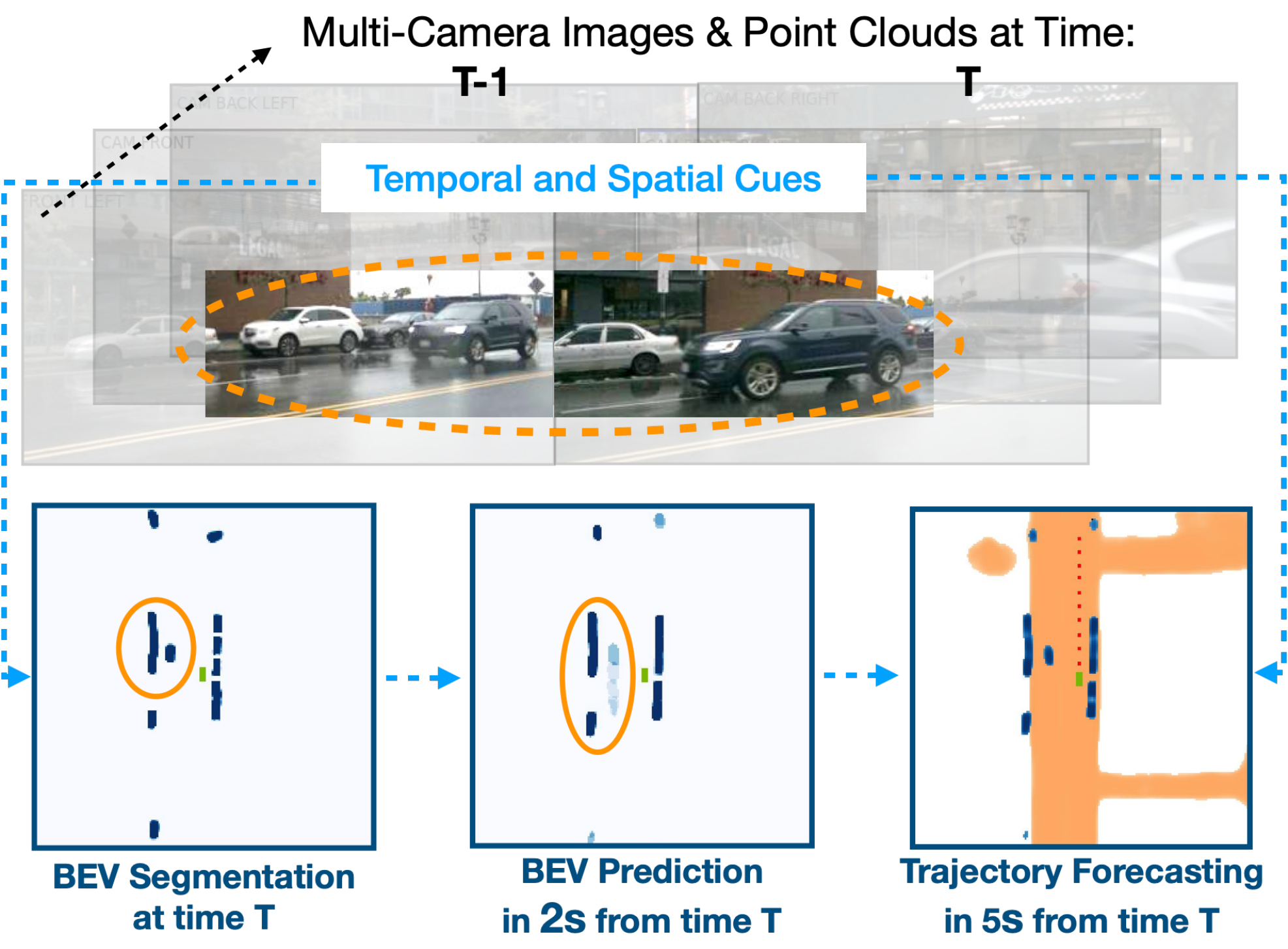}
   \caption{TLCFuse takes temporal input LiDAR and Camera data, accurately predicts vehicles' locations in the BEV at the reference time, even in highly occluded scenarios. It can also forecast the surrounding traffic scenes up to 2 seconds. As illustrated in the example, the white car visible at $T-1$ becomes occluded by the black SUV at $T$. TLCFuse captures the presence of the white car in its BEV segmentation at time $T$ (circled), and predicts the future trajectory of the moving black SUV (light blue trace within the circle). Leveraging these output BEV maps, TLCFuse forecasts the ego-vehicle's trajectory for the next 5 seconds.
   }
   
\label{fig:intro}
\vspace{-1.9em}
\end{figure}

Unlike previous studies that utilize temporal data solely for predicting future time steps, our work demonstrates that incorporating temporal cues can enhance how autonomous vehicles perceive their surroundings, particularly in complex scenarios involving object occlusion. 
We propose our innovative TLCFuse designed to process sequences of multi-modal sensor data inputs, capturing the temporal memory of the scene and objects to enhance occlusion-aware semantic BEV grid segmentation. By employing a fixed-size low-dimensional latent representation, our approach extracts spatial features from fused LiDAR and camera sensor readings at each time step throughout the temporal horizon. This representation serves as a stable spatio-temporal memory bank for feature extraction across time, ensuring coherence and efficiency in processing.
Moreover, TLCFuse is flexible to be applied to downstream tasks such as one-shot multi-step future BEV grids forecasting and ego-vehicle trajectory prediction.
In addition, TLCFuse is differentiable, allowing it to be integrated into an end-to-end training framework. One example for illustration can be seen in Fig. \ref{fig:intro}.

In summary, the contributions of our paper are: 
\begin{enumerate}
    \item \textbf{Novel Architecture:} TLCFuse integrates temporal multi-step spatial information derived from the fusion of LiDAR and camera sensor readings, to enhance occlusion-aware semantic BEV grid segmentation.
    \item \textbf{Accuracy:} Our experimental results show that TLCFuse outperforms existing state-of-the-art methods, especially when evaluating occluded and partially-occluded vehicles on the nuScenes dataset\cite{nuscenes}. 
    \item \textbf{Flexibility:} Experiments on downstream tasks are conducted by applying TLCFuse to predict multi-step future BEV grids in a one-shot manner and to forecast the ego-vehicle's trajectory.
    \item \textbf{Differentiability:} We offer insights on integrating TLCFuse within an end-to-end (E2E) trainable framework for full-stack driving tasks.

\end{enumerate}
\vspace{-0.3em}

%% file: sections/related_work.tex
\section{Related Work} \label{sec:related_work}

BEV perception methods have become popular in recent years in autonomous driving. Camera-based approaches try to find the correspondence between image pixels and cell locations in the BEV grids. LSS \cite{philion-lss} projects 2D features to 3D by inferring an implicit depth distribution over pixels.
Saha et al. \cite{imgs2map} use attention operations to associate image columns with their respective frustum projections in the BEV. PON \cite{roddick-pon} and VPN \cite{vpn} use MLPs to learn correspondences across multiple scales from the image space to the BEV. Learning these associations can be computationally costly and prone to overfitting. Methods like BEVFormer \cite{li2022bevformer} and CVT \cite{zhou2022cross} adopt cross-attention to find correspondences between images and BEV grids. LaRa \cite{bartoccioni2022lara} uses cross-attention to efficiently aggregate camera images and camera geometries in a encoder-decoder manner. BAEFormer\cite{pan2023baeformer} further improves the performance by cross-integrating multi-scale input image information. Nevertheless, camera-only methods lack real depth information for accurate results, while Lidar data can compensate for this. Lidar-based methods such as PillarSegNet \cite{pillarsegnet} associate the input data to each BEV cell using an orthographic projection, only taking into account the geometric structure of the scene. Yet, the sparse LiDAR points can't provide high-resolution information that is available in images. Consequently, the study of fusing multiple modalities becomes popular and promising. BEVFusion \cite{bevfusion} processes images and point clouds separately and uses the BEV as the shared space to fuse them together. Fishing Net \cite{hendy2020fishing} incorporates camera, radar and LiDAR in the input to predict future BEV semantic grids. TransFuseGrid \cite{salazar2022transfusegrid} also encodes images and point clouds separately into a BEV by fusing them with multi-scale self-attention transformers. LAPTNet \cite{diaz2023laptnet} projects the point cloud to the image plane to get depth information for the projection of image features to the BEV. Building upon these foundations, this paper introduces a novel approach that seamlessly integrates temporal sequences of multi-modal data processing. This integration not only facilitates the fusion of spatial information from LiDAR and camera sources but also enhances the incorporation of temporal cues, contributing to robust perception of the surrounding scene.

Following BEVdet4D\cite{huang2022bevdet4d} and SoloFusion\cite{park2022time}'s success in object detection by fusing temporal image features, BEVFormer \cite{bevformer} performs temporal self-attention with aligned BEV features to enhance perception results. FIERY \cite{hu2021fiery} performs temporal fusion of BEV features to predict vehicles' future states, became the first to combine perception and prediction in one network. Followed by BEVerse\cite{zhang2022beverse} that generates iterative flows for future state prediction and jointly reason object detection. Asghar et al.\cite{asghar2023vehicle} explored the integration of prior knowledge derived from DOGMs to predict agents' motion within BEV grids. Unlike these works, we leverage temporal information to tackle challenges in occlusion scenarios for semantic segmentation. Moreover, our approach can handle future scene forecasting and ego-car trajectory prediction.

%% file: sections/approach.tex
\section{Approach} \label{sec:approach}
\begin{figure*}[h]
	\centering
	    \includegraphics[trim={0 0 0 0},clip,width=1.9\columnwidth]{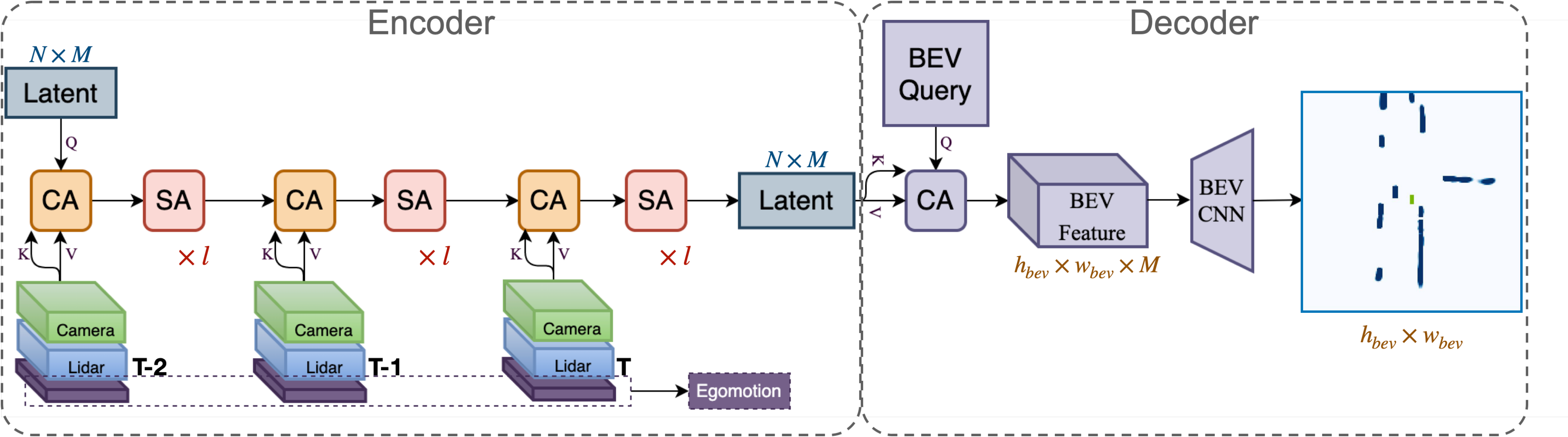}
   \caption{Overview of our proposed approach. At the encoding stage, a sequence of consecutive feature tensors at times $T-2$, $T-1$ and $T$ is input, where each feature tensor comprises concatenated domain-specific features from LiDAR, Camera and egomotion. A low-dimensional latent representation $L$ is utilized to extract spatio-temporal information from the input through a cross-attention (CA) layer and a few self-attention (SA) layers sequentially. At the decoding stage, a BEV query is employed to extract information from $L$, which is then stored in the BEV feature vector through a cross-attention operation. Subsequently, a CNN module is applied to refine the BEV feature into the target semantic BEV grid. }
\label{fig:overview}
\vspace{-1.6em}
\end{figure*}

In this section, we introduce TLCFuse, a novel architecture for occlusion-aware semantic BEV segmentation. TLCFuse is a transformer-based encoder-decoder network that leverages spatio-temporal information acquired through sequential multi-modal sensor fusion. 
Unlike existing methods that rely on either recurrent networks or external memory structures for capturing temporal dependencies, TLCFuse employs a temporal-augmented attention-based encoder to create a compact representation of input data over time. To the best of our knowledge, this paper is the first to propose such an architecture.
An overview of the pipeline is illustrated in Fig. \ref{fig:overview}. 

\subsection{Temporal Augmented Attention-based Encoder}\label{sec:approach-temp}

Attention-based transformer models have demonstrated their efficacy in capturing spatial relationships among multiple modalities\cite{jaegle2021perceiver,bevfusion,salazar2022transfusegrid,diaz2023laptnet}, providing accurate spatial cues for scene understanding. Nonetheless, there are known limitations hindering our integration of sequential sensor readings. Transformer models struggle with large inputs like point clouds due to their quadratic scaling with input size. Additionally, their fixed-length input sequences restrict them by a fixed-length memory to capture long-term dependencies for future predictions. Some existing models try to use recurrence or convolutional layers to explicitly model temporal dependencies from the input, or use external memory structures to store and retrieve long-term information\cite{hu2021fiery,bevformer,hu2023planning}. However, these techniques increase even more the computational cost. 

In response to the aforementioned limitations, we designed a novel attention-based encoder for TLCFuse, augmenting the comprehensive spatial feature extraction capabilities of transformers by integrating temporal information. We first initialize a fixed-sized low-dimensional latent-array $L\in\inR^{N\times M}$ with $N$ latent vectors of dimension $M$, and where $N$ is much smaller than the input feature size. This latent representation will function as a spatio-temporal memory bank, facilitating feature extraction across time through cross-attention operations. 
The encoder receives a temporal sequence of consecutive feature tensors as input, where each feature tensor comprises concatenated features from LiDAR, Camera and egomotion. At each time step, $L$ is utilized to capture intricate interactions across modalities and to extract spatial relationships among objects in the scene through a cross-attention layer, followed by a few self-attention layers. Subsequently, we proceed to the features of the next time step. The temporal information flow in the encoder mimics the progression of time in the real world, and such temporal awareness is crucial in maintaining a coherent understanding of object movements and relationships, addressing challenges associated with occlusions. 

\subsection{Multi-Modal Fusion}\label{sec:approach-fuse}

TLCFuse's encoder takes as its input sequential multi-modal fused feature tensors. At each time step, an input feature tensor comprises three geometrically informed modalities: multiple RGB camera frames, LiDAR point cloud, and the egomotion corresponding to the given reference time. We exploit the strong contextual ability of attention-based TLCFuse model to fuse these modalities in a trainable manner. 

Given a set of $K$ camera frames ${F_T^k\in\inR^{H \times W \times 3}}_{k=1}^K$ at each time $T$, we adopt a pretrained EfficientNet \cite{tan2019efficientnet} backbone to extract camera features $e_T^{cam}\in\inR^{K \times H/d_f \times W/d_f \times c}$ where $d_f$ is the down-sized factor and $c$ is the number of feature channels. 
Drawing inspiration from \cite{bartoccioni2022lara}, we initialize a set of positional embeddings which contain the intrinsic and the extrinsic information of the $k^{th}$ camera, and to transform the homogeneous coordinates of each feature pixel in the corresponding $k^{th}$ batch of $e_T^{cam}$, to project $e_T^{cam}$ from the image 2D coordinate system to the 3D ego-frame coordinate system. Then, we concatenate the $K$ transformed embeddings and pass it through a 2-layer MLP module \cite{hendrycks2016gaussian} to yield the final positional embedding $\O_T^{cam}\in\inR^{K \times H/d_f \times W/d_f \times c}$, and to project $e_T^{cam}$ to the BEV coordinates. 

Given the point cloud $P_t\in\inR^{4\times D}$ with $D$ points taken from a LiDAR sensor on the ego-vehicle, we adopt a PointPillar \cite{pointpillars} backbone to group the points in the 3D space into a grid of 'pillars' in the BEV plane. Within each pillar, the information of all the points that fall inside are combined to yield the LiDAR feature map $e_T^{pc}\in\inR^{h \times w \times c}$ where $(h,w)$ is the spatial shape of the BEV plane and $c$ is the same number of feature channels as used in the camera features $e_T^{cam}$. Additionally, we concatenate a Fourier positional embedding $\O_T^{pc}\in\inR^{h\times w\times c}$ to the feature tensor $e_T^{pc}$, to strengthen the geometric alignment between the 3D point cloud coordinates and the BEV coordinates.

On top of these modalities, we incorporate the egomotion embedding $\O^{ego}_{T-h\rightarrow T}\in\inR^{6\times c} , h = 1,...,H$ into our input feature tensor at each time step, where $H$ is the temporal horizon. The egomotion represents the rotation and translation of the ego-vehicle from a past time step $T-h$ to the reference time $T$. For time $T$, the egomotion embedding $\O^{ego}_T$ is a zero-filled tensor. As the ego position of the autonomous vehicle evolves over time, egomotion embedding plays a crucial role in aiding TLCFuse's temporal mechanism to dynamically adapt to changes in perspective. 

\subsection{TLCFuse's Decoder}\label{sec:approach-decoder}

TLCFuse's decoder consists of a BEV query and a BEV refinement network based on ResNet18 \cite{he2016deep}. The BEV query $Q\in\inR^{h_{bev}\times w_{bev}}\times c_{bev}$ is initialized to the spacial size of the target output BEV grid. At the decoding stage, $Q$ attends to the latent representation $L$ through a cross-attention, and passes the obtained BEV feature map $\inR^{h_{bev} \times w_{bev} \times M}$ to the CNN module to be decoded into the output grid. Simply by extending the output BEV channel to $\inR^{h_{bev} \times w_{bev} \times f}$, we can obtain a sequence of BEV predictions from time $T$ to $T+f$ in a one-shot manner. 

\subsection{Trajectory Forecasting with TLCFuse}\label{sec:approach-plan}
\begin{figure}[h]

	\centering
	    \includegraphics[trim={0 0 0 0},clip,width=1\columnwidth]{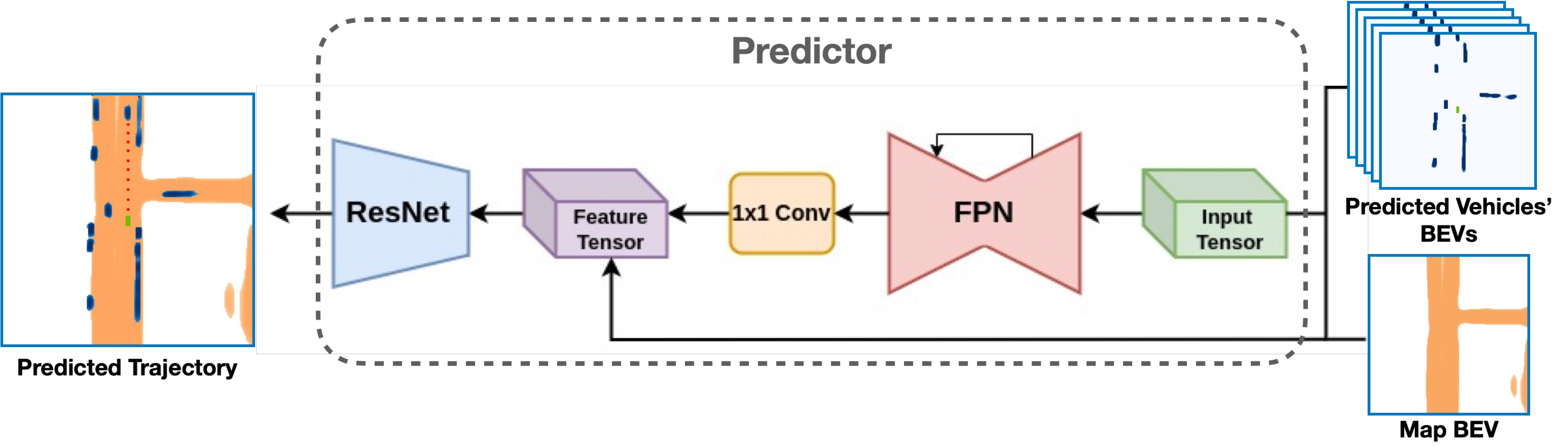}
   \caption{We introduce a trajectory forecasting network designed to anticipate the future trajectory of the ego-vehicle. 5 BEV grid predictions of the Vehicle and 1 BEV grid of the Drivable Area serve as inputs to the predictor. The network analyzes these inputs to discern the most probable route for the ego-vehicle over the next 5 seconds.}
\label{fig:planner}
\vspace{-1.6em}
\end{figure}

TLCFuse can be applied to downstream task to anticipate the future trajectory of the ego-vehicle within the scene. As shown in Fig. \ref{fig:planner}, we utilize TLCFuse to generate 5 BEV predictions for the vehicles and 1 BEV grid for the Drivable Area. We concatenate these grids as the input to a Feature Pyramid Network (FPN)\cite{lin2017feature} followed by a $1\times1$ convolution layer. The extracted feature receives a skip connection from the input Drivable Area map to boost the memory of the driving lane. Finally, a ResNet \cite{he2016deep} is adopted to predict the final trajectory and heading $\inR^{p \times 3}$ of the ego-vehicle, where $p$ is the prediction time horizon. In this paper, we adopt pretrained TLCFuse model in generating inputs for the Predictor. However, ongoing work is underway towards end-to-end trainable TLCfuse for motion planning to be discussed in future works.

%% file: sections/experiments.tex
\begin{figure*}[ht!]
 \centering
  
    \includegraphics[trim={0 0 0 0},clip,width=1.85\columnwidth]{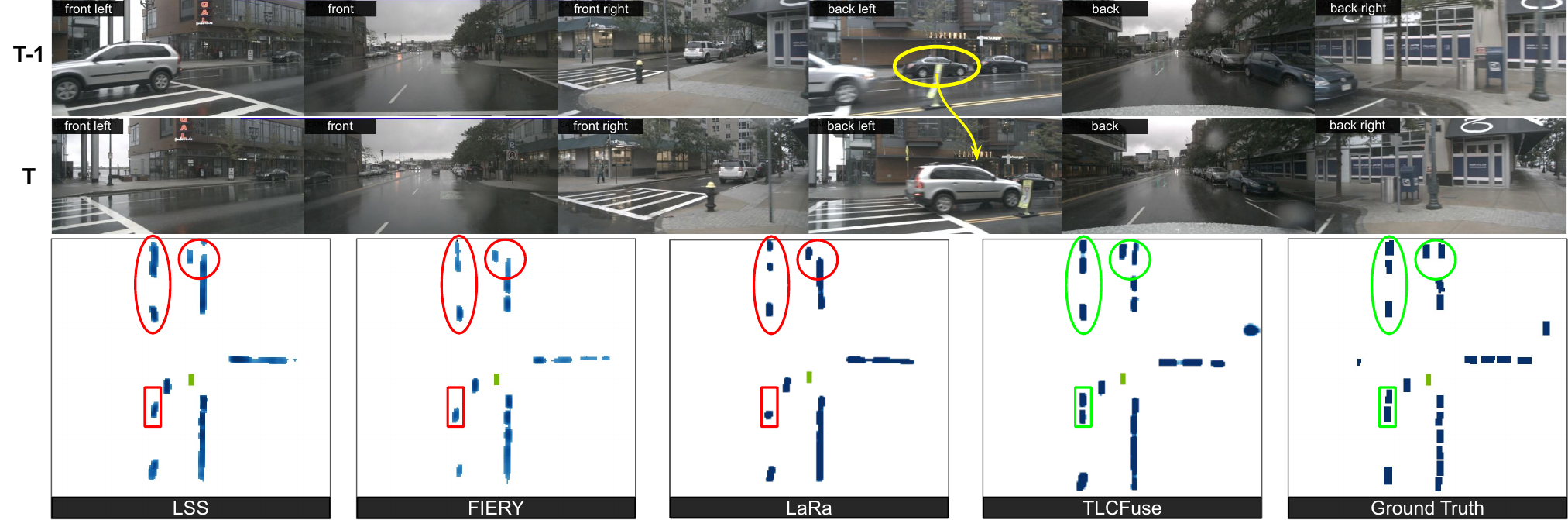}  

   \caption{Qualitative example of occlusion-aware BEV segmentation. The back-left camera of the ego-vehicle records two black cars being occluded by a silver car. Prior works LSS \cite{philion-lss}, FIERY \cite{hu2021fiery} and LaRa \cite{bartoccioni2022lara} fail to generate the locations of the occluded cars in their semantic maps (indicated by a red box). While TLCFuse correctly locates the occluded vehicles (indicated by a green box). Additionally, TLCFuse generate the clearest and sharpest semantic map than the others (see the areas indicated by green circles). }
\label{fig:qualitative}
\vspace{-1.4em}
\end{figure*}

\section{Experiments} 

\subsection{Dataset}
We use nuScenes \cite{nuscenes} and nuPlan \cite{nuplan} datasets to evaluate our proposed framework. NuScenes consists of 1000 scenes captured from different locations and times of the day; each scene's duration is approximately 20 seconds providing data from an entire sensor suite, including LiDAR and cameras. The 6 cameras provide a $360^{\circ}$ view field around the ego-vehicle with overlaps. The synchronized keyframes(images, LiDAR) are annotated at 2Hz and all objects come with attributes such as visibility level in the camera images. 
NuPlan provides more than 1200 hours of realistic human driving data with distinct and diverse traffic behaviors. It is the world's first large-scale standardized dataset for autonomous vehicle motion planning.
 
We use the nuScenes dataset to train and test TLCFuse for semantic grid prediction tasks. For trajectory forecasting, we first train and test our motion predictor on the nuPlan dataset, then evaluate the full pipeline end-to-end on the nuScenes validation set.
\vspace{-0.4em}

\subsection{Metrics}

For semantic grid predcition tasks, we use Intersection Over Union (IoU) to evaluate our model's performance. IoU is used to quantify the amount of overlapping between the generated semantic grids and the corresponding ground truth. To better show TLCFuse's performance under different occluded scenarios, we incorporate a set of visibility levels of the objects provided by the nuScenes dataset while computing IoUs for comparison. The visibility percentage specifies the fraction of visible pixels for an object over the whole camera rig. For nuScenes, the visibility levels are organized in 4 bins: 0-40\%, 40-60\%, 60-80\%, 80-100\%. 


\subsection{Implementation Details} \label{sec:implementation}

We analyze sequences of input data with the temporal horizon extending over 1 second, covering three consecutive time steps from the past, aligning with the 2Hz frequency of the nuScenes dataset. At each time step, the input modalities include 6 RGB camera images with sizes scaled to $224\times480$ and 10 sweeps of LiDAR point cloud data.

In the encoder, the latent-array $L$ is initialized to the size of $256 \times 256$ with zero mean, standard deviation 0.02 and value limits of (-2, 2). In the decoder, we initialize the BEV Query to the size of $200 \times 200 \times 256$. The target output BEV grid is $200\times200$ in cells, with a resolution of 50$cm$ per cell, corresponding to a $100\times100$ meters area centered around the ego-vehicle. When predicting the future BEV grids, we set prediction horizon $f = 5$, corresponding to a 2 seconds into the future. We train TLCFuse on nuScenes training set for 25 epochs with a batch of 2, and test on the nuScenes validation set. In motion forecasting experiment, we train the proposed motion forecasting network independently on the nuPlan dataset for 76 epochs. Subsequently, we conduct an evaluation on the nuPlan validation set using TLCFuse without fine-tuning. This involves generating 5 BEV grid predictions for the vehicles and 1 BEV grid for the drivable area. These grids are then input into the motion planner network to forecast the trajectory of the ego-vehicle for 5 seconds.

All our experiments are conducted in a computer with 8 GPUs Nvidia Tesla V100.

\vspace{-0.3em}

%% file: sections/results.tex
\section{Results} \label{sec:result}
\subsection{Qualitative Results}

Qualitative results in Fig. \ref{fig:qualitative} illustrate TLCFuse's performance in occluded scenarios. In this instance, the scene captured by the back-left camera of the ego-vehicle reveals that a black car (highlighted by the yellow circle) parked alongside the road becomes concealed behind a moving silver SUV at time $T$
Comparing the BEV segmentation results generated by LSS \cite{philion-lss}, FIERY \cite{hu2021fiery}, LaRa \cite{bartoccioni2022lara}, and TLCFuse, our method stands out as the only one capable of accurately and clearly recovering the occluded black car in its generated BEV map (indicated by the green box).
In addition, TLCFuse also excels at perceiving objects located far from the ego-vehicle and accurately mapping them onto the Bird's-Eye View (BEV) grids (as evident within the green circles in TLCFuse's result). From this figure, it is apparent that TLCFuse's BEV semantic segmentation maps align more closely with the actual ground truth than those of other state-of-the-art methods.

\subsection{Quantitative Results}

In Tables \ref{tab:vis123} and \ref{tab:vis1}, we present quantitative results of TLCFuse in the BEV segmentation task, comparing it to previous works using the nuScenes dataset. For BEV grid segmentation, nuScenes provides multiple semantic categories, including Vehicle, Drivable Area, Human, and Walkway.
Not all existing BEV segmentation approaches operate on the same categories. In Table \ref{tab:vis123}, we present the IoU scores specifically computed for the Vehicle category. Our scores are reported across three different visibility levels provided by nuScenes.
Visibility greater than 0\% (second column in the table) represents all the vehicles within the scenes. Visibility greater than 40\% (third column in the table) means that only the visible vehicles (not significantly occluded) in the scenes are considered. And visibility between 0\% and 40\% (fourth column in the table) only consists of partially or completely occluded vehicles in the scenes. 
We compare to the state-of-the-art methods that produce BEV segmentation results for the Vehicle category. We observe that TLCFuse consistently achieves the best scores compared to others across all different visibility levels. Our outstanding performance under visibility levels between 0\% and 40\% further proves that TLCFuse can provide robust perception under occluded situations.

\begin{table}[t]
\begin{adjustbox}{width=\columnwidth,center}
\centering
\begin{tabular}{l|c|c|c|c|c}
\hline
Method  & Modalities                  & vis$\geq$0\%       & vis$>$40\%       & 0\%$<$vis$<$40\% & FPS \\ \hline
LSS \cite{philion-lss} & C & 32.1 & 34.81 & 8.65 & 25 \\
BEVFormer\cite{li2022bevformer} & C & 43.2 & - & - & 2 \\
CVT \cite{cross-view-transformers} & C & - & 36.0 & - & 48\\
FIERY Static\cite{hu2021fiery} & C & 35.8 & 39.26 & 8.54 & 8 \\
LaRa\cite{bartoccioni2022lara}   & C          &     35.81            &     37.92             &      9.41           & -            \\ 
BAEFormer\cite{pan2023baeformer} & C & 36.0 & 38.9 & - & 45\\
TFGrid  \cite{salazar2022transfusegrid} & C+L      & 35.88   &- &- &18.3       \\
LAPTNet  \cite{diaz2023laptnet}  & C+L    & 40.13   & - & - & 43.8        \\
\rowcolor[HTML]{EFEFEF}
\textbf{TLCFuse} & C+L+T&    \textbf{43.65}   &     \textbf{46.40}    &      \textbf{13.34}    & 9          \\ \hline
\end{tabular}
\end{adjustbox}
\caption{\textbf{Vehicle BEV segmentation with different visibility levels.} TLCFuse was trained on nuScenes with all vehicles within the scenes, and different visibility levels are only considered for validation. }
\label{tab:vis123}
\vspace{-1.2em}
\end{table}


\begin{figure*}[!htbp]
    \centering
    \begin{subfigure}[t]{0.45\textwidth}
        \centering
        \includegraphics[trim={0 6.5 0 0},clip,width=1.3\columnwidth]{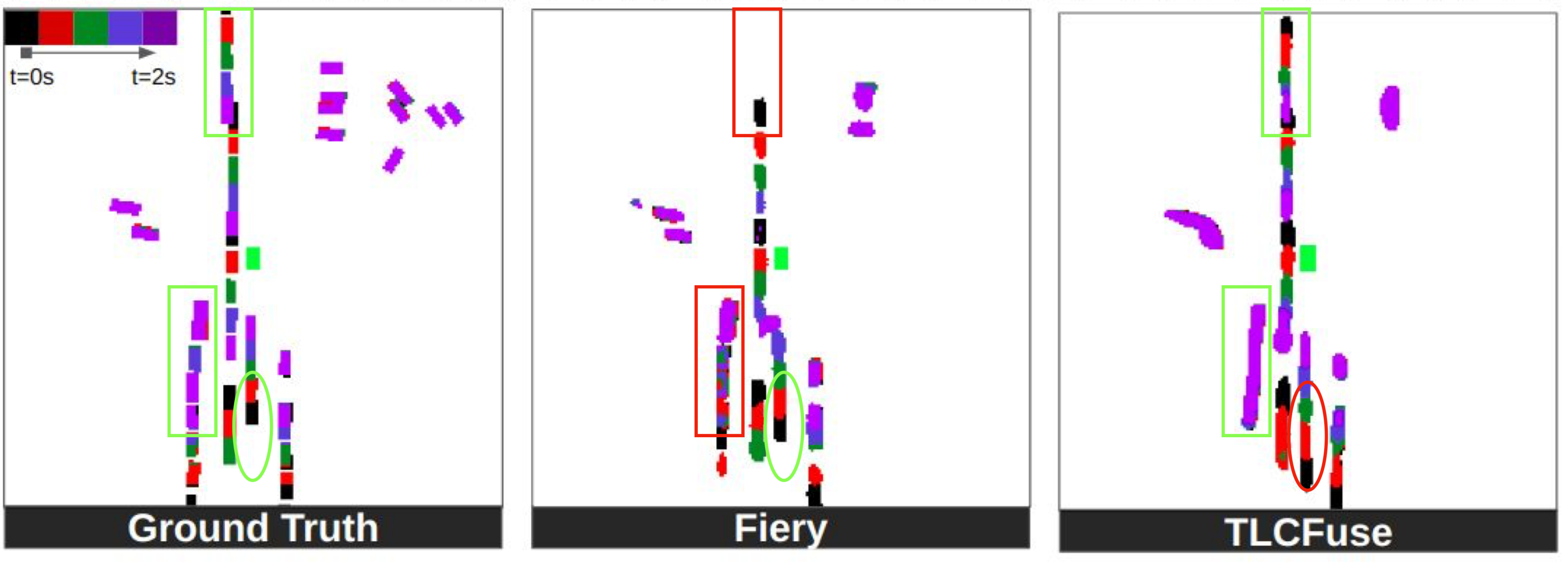}
        \caption{}
        \label{fig:bev_pred}
    \end{subfigure}
    \hfill
    \begin{subfigure}[t]{0.45\textwidth}
        \centering
        \includegraphics[trim={0 0 0 0},clip,width=0.84\columnwidth]{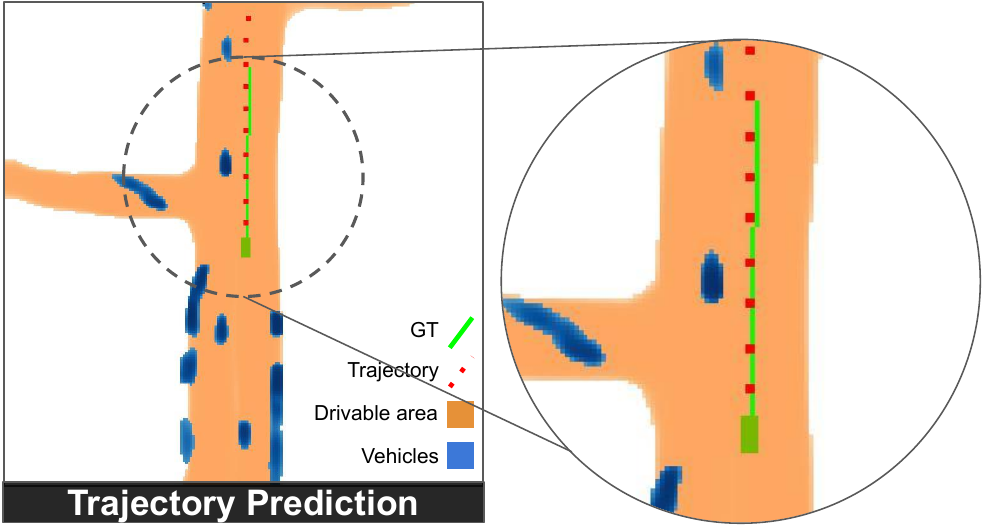}
        \caption{}
        \label{fig:traj_pred}
    \end{subfigure}
    \caption{(a) Qualitative results of one-shot multi-step future BEV grid prediction on the nuScenes dataset are presented. We compare TLCFuse's results against the state-of-the-art model Fiery\cite{hu2021fiery} and the ground truth. The predictions are color-coded for 2 seconds into the future, with green boxes and circles indicating accurate predictions compared to the ground truth, and red ones representing mistakes. TLCFuse produces qualitatively satisfying predictions, showcasing its effective performance despite its naive prediction mechanism. (b) Qualitative results of trajectory forecasting on nuScenes dataset. The predicted 5seconds trajectory of the ego-vehicle is shown in red dots overlayed onto the semantic maps of the drivable area and surrounding vehicles at time 0s. The groundtruth future trajectory is drew in a green line. We see that our model predicts reasonable and accurate trajectory for the ego-vehicle. }
    \label{fig:combined}
\vspace{-1.4em}    
\end{figure*}

\begin{table}[t]
\begin{adjustbox}{width=\columnwidth,center}
\centering

\begin{tabular}{@{}l|c|ccc|c@{}}
\toprule
Method                                       & Modalities & Human  & Drivable Area  & Walkway  & FPS   \\ \midrule
Pyramid Occupancy Network \cite{roddick-pon} & C          & 8.2    & 60.4           & 31.0     & 22.3  \\
LSS  \cite{philion-lss}                      & C          & 9.99   & 72.9           & 51.03    & 25  \\
M2BEV  \cite{m2bev}                          & C          & -      & 75.9           & -        & 4.3   \\
Translating Images into Maps \cite{imgs2map} & C          & 8.7    & 72.6           & 32.4     & -     \\
BEVFormer \cite{bevformer}                   & C          & -      & 77.5           & -        & 2    \\
UniAD \cite{hu2023planning}                  & C+T        & -      & 69.1            &-        & 2
\\
\midrule
PointPillars baseline  \cite{pointpillars}   & L          & 0.0    & 58.44          & 33.66    & -     \\
TFGrid     \cite{salazar2022transfusegrid}   & C+L        & -      & 78.87          & 50.98    & 18.3 \\  
BEVFusion    \cite{bevfusion}                & C+L        & -      & \textbf{85.5}  & \textbf{67.6}              & -      \\ 
LAPTNet    \cite{diaz2023laptnet}            & C+L        & 13.8   & 79.43          & 57.25    & \textbf{43.8} \\

\rowcolor[HTML]{EFEFEF}
TLCFuse                                      & C+L+T      & \textbf{15.24}  & 76.79 & 46.37 & 9 \\
\bottomrule

\end{tabular}
\end{adjustbox}
\caption{ \textbf{BEV segmentation qualitative results on nuScenes dataset for other semantic classes}: Human, Drivable Area and Walkway. TLCFuse outperforms other models in the Human category and performs comparably with state-of-the-art models in the Drivable Area and Walkway categories.}
\label{tab:vis1}
\vspace{-1.8em}
\end{table}

We transition from evaluating the vehicle category to assessing other categories. Specifically, we focus on three of the most commonly adopted categories: Human, Drivable Area, and Walkway. It is important to note the absence of a visibility mask for the Drivable Area and Walkway categories in nuScenes. Unlike vehicle category, drivable areas, and walkways are typically considered accessible and unobstructed for vehicle navigation. Therefore, we generate Table \ref{tab:vis1} without any visibility mask and compare it to state-of-the-art methods that include these three categories. From the table, we observe that TLCFuse performs admirably across the Drivable Area and Walkway categories and achieves the best results for the Human category.
Additionally, we would like to note that the state-of-art model BEVFusion\cite{bevfusion} achieves the best scores in Drivable Area and Walkway categories by processing higher-resolution camera images and focusing specific on road map generation. We would like to adopt the same strategy in TLCFuse in our future work.

\subsection{Ablation Study}

We conducted an ablation study to elucidate the individual contributions of different input modalities, with the results presented in Table \ref{tab:ablation}. This table provides insights into the impact of varying input modalities on the performance of our proposed network.

\begin{table}[h]
\begin{adjustbox}{width=0.65\columnwidth,center}
\fontsize{8}{10}\selectfont
\centering
\begin{tabular}{c|ccc|c}
\hline
\multicolumn{1}{c|}{\multirow{2}{*}{ID}}  & \multicolumn{3}{c|}{Modality} & \multicolumn{1}{c}{\multirow{2}{*}{IoU}} \\
\multicolumn{1}{c|}{}                     & Camera   & LiDAR   & Temporal &  \multicolumn{1}{c}{} \\  \hline
1                                         & \checkmark   &     &          &  35.61  \\
2                                         &        &\checkmark &          & 31.09   \\
3                                         &\checkmark  & \checkmark &     & 42.56    \\
4                                         & \checkmark &       &\checkmark& 35.67  \\ 
5                                         &  & \checkmark      &\checkmark& 32.01    \\  
6                                         & \checkmark & \checkmark  &\checkmark  & \textbf{43.65}      \\ \hline      
\end{tabular}
\end{adjustbox}
\caption{\textbf{Ablation on different modalities.} We show here the BEV segmentation IoU scores using different combinations of input modalities.}
\label{tab:ablation}
\vspace{-1.4em}
\end{table}

In this table, we provide IoU scores for BEV grid segmentation on the nuScenes dataset using all the vehicles in the scene. We compare the performances given by different modality combinations: 1) only using multiple camera images; 2) only using LiDAR point cloud; 3) using camera and LiDAR fusion; 4) using a sequence of camera images; 5) using a sequence of LiDAR data; and 6) using temporally fused camera and LiDAR data. From the table, the best score was achieved by fusing temporal LiDAR and camera data, namely TLCFuse. While the fusion of LiDAR and camera already improves performance compared to using a single modality, adding temporal information further boosts the performance.

\subsection{Motion Prediction}

We applied TLCFuse to two downstream motion prediction tasks: one-shot multi-step BEV grid prediction and trajectory forecasting for the ego-vehicle. These tasks not only validate the effectiveness of our model but also highlight its flexible applicability across a wide range of autonomous driving scenarios.

We first employed TLCFuse in the BEV grid prediction task, as explained in Section \ref{sec:approach-decoder}. Existing motion prediction approaches, such as FIERY\cite{hu2021fiery} and BEVerse\cite{zhang2022beverse}, often rely on recurrent networks (RNNs, LSTMs) to predict accurate future instances \textbf{in an iterative} fashion. However, these recurrent networks may increase the network's complexity and inference latency. In contrast, TLCFuse can be used to generate multiple predictions \textbf{in one shot} by simply adjusting its decoder stage's output channel. This allows TLCFuse to achieve a 9 FPS inference latency, which is much faster than FIERY\cite{hu2021fiery} and BEVerse\cite{zhang2022beverse}, with an average speed of 2 FPS.

We present a qualitative example in Fig. \ref{fig:bev_pred}. In this illustration, multi-step BEV predictions are generated in one shot. The predictions are color-coded to visualize the time range from 0s to 2s, progressing from black to purple, and are overlaid on the map of the Drivable Area generated by TLCFuse. When compared to the ground truth and the state-of-the-art model Fiery\cite{hu2021fiery} on its left, TLCFuse yields qualitatively satisfying predictions. In particular, TLCFuse provides more accurate predictions for both moving and static objects in the scene than Fiery, as indicated in the green/red boxes in the figure. However, as circled in the figure, TLCFuse also makes mistakes in predicting the accurate locations of this moving vehicle.

The second downstream task we undertake involves predicting the future trajectory of the ego-vehicle. We have detailed the network and experiment designs in Sections \ref{sec:approach-plan} and \ref{sec:implementation}. We pursued this task due to its significant relevance in showcasing our model's capacity to comprehend surrounding information and guide the ego-vehicle effectively. Furthermore, this task aligns with our overarching research goal of transitioning our approach towards an end-to-end framework.

One qualitative example is shown in Fig. \ref{fig:traj_pred}. Our model predicted the trajectory for the ego-vehicle 5seconds into the future as indicated by the red dots. To enhance result visualization, we drew the groundtruth future trajectory in a green line, and overlaid the predicted trajectory on the drivable area map and surrounding vehicles at time 0s. The model reasonably predicts the ego-vehicle following the current lane on the road with a constant speed.

For more experimental results, please refer to  \url{https://github.com/gsg213/TLCFuse}.

%% file: sections/conclusion.tex
\section{Conclusions and Future Work}\label{sec:conclusion}
In this paper, we present TLCFuse. A novel, flexible and differentiable architecture that extract spatio-temporal cues for occlusion-aware BEV segmentation. We conducted extensive experiments on the nuScenes datase, and our experimental results show that TLCFuse outperforms existing methods, especially under occluded scenarios. In addition, we also demonstrate the flexibility of TLCFuse by applying it to two downstream tasks: multi-step BEV prediction in a one-shot manner and trajectory forecasting of the ego-vehicle. In future work, we will continue the ongoing work of a new end-to-end trainable pipeline integrating TLCFuse and the proposed motion planner network. 

%% file: sections/acknowledgement.tex
\section{Acknowledgment} \label{sec:acknowledgment}

Experiments presented in this paper were carried out using the Grid'5000 testbed, supported by a scientific interest group hosted by Inria and including CNRS, RENATER and several Universities as well as other organizations (see https://www.grid5000.fr).